\documentclass[final,3p]{CSP}
\usepackage{url}
\usepackage{amssymb}
\usepackage{changepage}
\usepackage{subfigure}
\usepackage{caption}
\usepackage[section]{placeins}
\usepackage{verbatim}
\usepackage{float}
\usepackage{color}
\usepackage{multirow}
\usepackage{algorithm}  
\usepackage{algorithmicx}  
\usepackage{algpseudocode}  
\usepackage{amsmath}
\usepackage{import}

\begin{document}

\begin{frontmatter}

\title{SolarNet: A Deep Learning Framework to Map Solar Power Plants In China From Satellite Imagery}

\author{Xin Hou}
\author{Biao Wang}
\author{Wanqi Hu}
\author{Lei Yin}
\author{Haishan Wu\corref{correspondingauthor}}
\cortext[correspondingauthor]{Corresponding author}
\ead{haishanwu@webank.com}

%\author[mysecondaryaddress]{Global Customer Service\corref{mycorrespondingauthor}}
%\cortext[mycorrespondingauthor]{Haishan Wu}

\address{WeBank AI Group}

\begin{abstract}\rm

	\begin{adjustwidth}{2cm}{2cm}{\itshape\textbf{Abstract:}} 
	Renewable energy such as solar power is critical to fight the ever more serious climate change. China is the world leading installer of solar panel and numerous solar plants were built. In this paper, we proposed a deep learning framework named SolarNet which is designed to perform semantic segmentation on large scale satellite imagery data to detect solar farms. SolarNet has successfully mapped 439 solar farms in China, covering near 2000 square kilometers, equivalent to the size of whole Shenzhen city or two and a half of New York city. To the best of our knowledge, it is the first time that we used deep learning to reveal the locations and sizes of solar farms in China, which could provide insights for solar power companies, market analysts and government
	\end{adjustwidth}
	
	\begin{adjustwidth}{2cm}{2cm}{\itshape\textbf{Keyword:}} 
		Deep Learning, Semantic Segmentation, Satellite Imagery, Renewable Energy, Solar Power
	\end{adjustwidth}
\end{abstract}

\end{frontmatter}

\section{Introduction}
While climate change has become one of the greatest threats to our world, renewable energy such as solar power is critical to fight climate change\cite{chu2012opportunities,agnew2015effect}. China, as the world's leading installer of solar photovoltaics (PV), is the world's largest producer of solar PV power and massive solar farms were built not only to produce clean energy but also to reduce poverty. 

However, one question remains to be answered: where are those solar farms located?  Mapping the location of solar farms and tracking its installation progress is particularly important for the following aspects: first, it allows the government to gauge the development of solar power industry and make strategies; second, it helps the solar power company to quantify and optimize the efficiency of solar panels; third, it is useful for investors to evaluate the operation of solar power companies. Obviously, it is impractical to locate solar farms on maps manually. What if we can trace them systematically from the sky? Most recently, more and more companies have launched satellites into space, produced massive satellite imagery data and therefore accelerated its commercialization in various fields.  

In this paper, we proposed a deep learning framework named \textbf{SolarNet}, which is used to analyze large-scale high-resolution satellite imagery data and is able to accurately identify hundreds  visible large solar farms in China while many of those are built in deserts, mountains and even lakes.  To the best of our knowledge, it is the first time that the locations and sizes of solar farms in China are tracked  by mining satellite imagery data through deep learning algorithms.

\section{Related Works}
In this section, we give a brief review of related works. Semantic segmentation\cite{long2015fully} is an important computer vision technique that has been widely  applied to detect objects from remote sensing imagery data, such as urban architectural segmentation\cite{wei2004urban,bischke2019multi}, road extraction\cite{mokhtarzade2007road}, crop segmentation\cite{rydberg2001integrated}, etc.  However, compared with natural images,  segmentation on satellite imagery data is much more challenging due to: 1) the resolution of different satellites may be not consistent, 2) the size of satellite is huge which may lead to huge computational cost, 3) the background, cloud, reflection of sunshine etc. could also complicate the segmentation, 4)the texture of solar panels may also vary due to various sensor specs.  Our framework SolarNet  which could detect solar farms from satellite imagery data is designed based on semantic segmentation. 

\noindent\textbf{Semantic Segmentation:}
Deep learning has achieved great success in semantic segmentation task\cite{krizhevsky2012imagenet}. In 2014,  Full Convolutional Network (FCN)\cite{long2015fully}, which replaced the network's fully connected layer with convolution, was proposed and achieved much higher accuracy than the patch classification method\cite{varma2008statistical}. Recently, \cite{li2019expectation} proposed by Xia Li on ICCV 2019 demonstrated a state-of-the-art segmentation algorithm named EmaNet.

\begin{figure}[!htbp]
	\begin{center}
		\includegraphics[scale=.65]{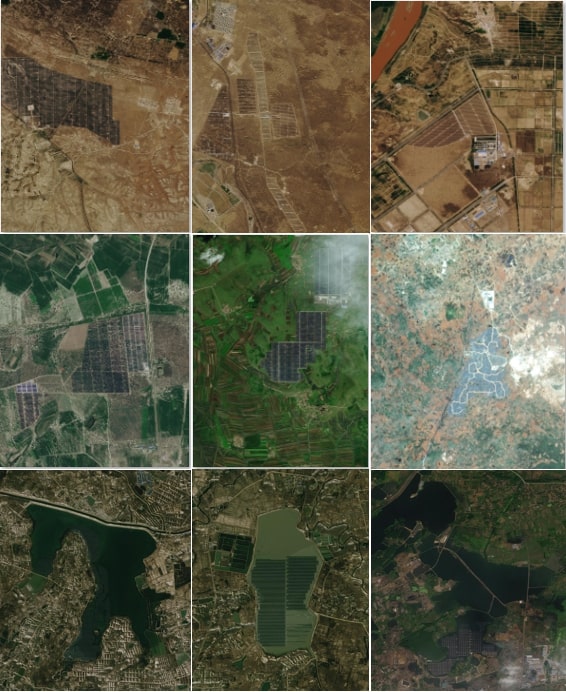}
		\caption{Part of solar farms in China. The first row shows solar power plants in the deserts, the second rows shows solar power plants in the mountains while the last row shows solar power plants in the lakes. One can see the complex backgrounds in those images. }
		\label{fig_dataset}
	\end{center}
\end{figure}

\noindent\textbf{Solar Panel Detection:} 
Most recently, Yu etc.\cite{yu2018deepsolar} proposed a framework called DeepSolar which successfully located  the civil solar panels in the United States and developed a public data set. Their data set mainly focused on household solar power planes in the US, by contrast, most of the large solar power plants in China were built in the fields with complex background such as deserts, mountains and even lakes as shown in Figure \ref{fig_dataset}, which pose more challenges to the detection task.  Our algorithm addressed those difficulties by combining the advantage of  FCN and EmaNet. In order to fully evaluate the proposed segmentation method, we also particularly created a  satellite imagery data set of the solar plants in China to train our model. 

\section{Method}

SolarNet  is based on Expectation-Maximization Attention Networks (EMANet). In order to compare the performance, we used UNet as a baseline algorithm, which is one of most popular deep learning based semantic segmentation methods.  

\subsection{UNet }

\begin{figure*}[htb]
	\centering{\includegraphics[scale=0.35]{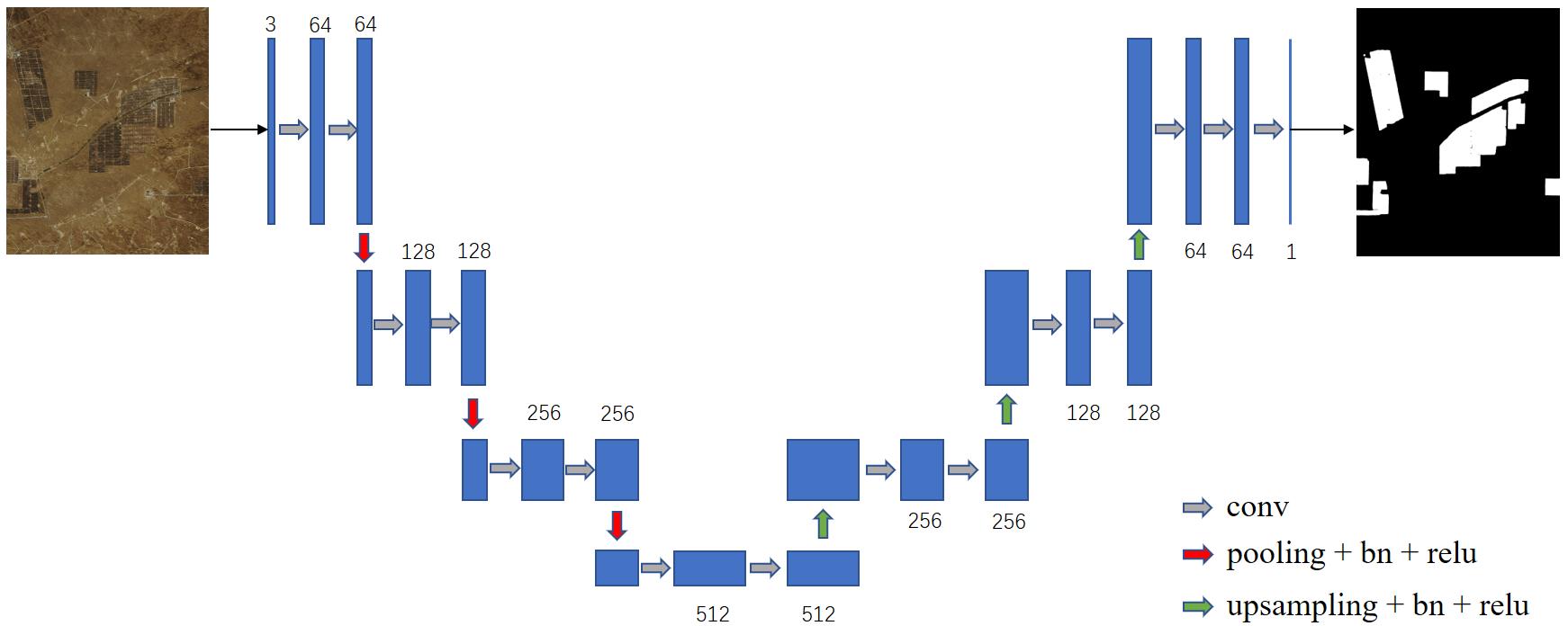}}
	\caption{UNet Architecture.}
	\label{fig:VF_CVF}
\end{figure*}

Different from the classic Convolutional Neural Networks (CNN), the  convolutional layer of FCN adopts the fully connected layer to obtain fixed-length feature vectors for classification\cite{dai2016r}, and thus is able to deal with input images with any size.  The deconvolution layer of FCN performs the feature map of the last volume-bases layer. This architecture can produce a prediction for each pixel, while retaining the spatial information in the original input image. The UNet architecture which stems from FCN was first proposed by [] is used as a baseline model and the net architecture is illustrated in Figure \ref{fig:VF_CVF}.

The network architecture is described in detail in Table \ref{tab:unet_detail}. It has tow parts: a contracting path and an expansive path. The contracting path follows the typical architecture of a convolutional network. we uses two repeated convolutions with 3$\times $3 kernerl size, while each is followed by a batch normalization layer and a rectified linear unit, a 2$\times $2 max pooling operation with stride 2 for downsampling. At each downsampling step we made the number of feature channels becomes to double times. In the expansive process every step consists of upsampling feature map followed by a 2$\times $2 convolution that halves the number of feature channels, a concatenation with the correspondingly cropped feature map from the contracting path, and two 3$\times $3 convolutions, each followed by a BN layer and a ReLU layer. In the final layer, a 1$\times $1 convolution is used to map each 2-component feature vector to the desired number of classes  whether this pixel is solar plane or not. The network has 17 convolutional layers in total.

% \begin{figure*}[hbp]
% \begin{center}
% \includegraphics[scale=0.5]{fig/Unet.png}
% \caption{UNET Architecture.}
% \end{center}\label{fig:VF_CVF}
% \end{figure*}

%\begin{comment}
\begin{table}[H]
	\centering
	\begin{tabular}{cccc}
		\hline
		\multicolumn{4}{c}{INPUT}\\
		\hline
		3x3 conv 64 dim$\rightarrow$ & 3x3 conv 64 dim$\rightarrow$ & pooling $\rightarrow$& BN \& RELU\\
		\hline
		3x3 conv 128 dim$\rightarrow$& 3x3 conv 128 dim$\rightarrow$& pooling $\rightarrow$& BN \& RELU\\
		\hline
		3x3 conv 256 dim$\rightarrow$& 3x3 conv 256 dim$\rightarrow$& pooling $\rightarrow$& BN \& RELU\\
		\hline
		3x3 conv 512 dim$\rightarrow$& 3x3 conv 512 dim$\rightarrow$& pooling $\rightarrow$& BN \& RELU\\
		\hline
		3x3 conv 512 dim$\rightarrow$& 3x3 conv 512 dim$\rightarrow$& upsampling $\rightarrow$& BN \& RELU\\
		\hline
		3x3 conv 256 dim$\rightarrow$& 3x3 conv 256 dim$\rightarrow$& upsampling $\rightarrow$& BN \& RELU\\
		\hline
		3x3 conv 128 dim$\rightarrow$& 3x3 conv 128 dim$\rightarrow$& upsampling $\rightarrow$& BN \& RELU\\
		\hline
		3x3 conv 64 dim$\rightarrow$& 3x3 conv 64 dim$\rightarrow$& upsampling $\rightarrow$& BN \& RELU\\
		\hline
		\multicolumn{2}{c}{1x1 conv 2 dim}$\rightarrow$& \multicolumn{2}{c}{SoftMax}\\
		\hline
	\end{tabular}
	\caption{UNet architecture detail}
	\label{tab:unet_detail}
\end{table}
%\end{comment}

%\subsection{Expectation-Maximization Attention Networks for Semantic Segmentation}
\subsection{SolarNet: a multitask Expectation-Maximization Attention Networks}

Attention mechanism have been widely used for various tasks. The proposed Expectation-Maximization Attention (EMA) module \cite{moon1996expectation} is robust with regard to the variance of input and is also efficient in terms of memory and computational power\cite{wang2018non}. For a simple introudction, we consider an input feature map $X$ of size $C\times H \times W$ from a single image. $X$ was the intermediate activated feature map of a CNN. We reshaped $X$ into $N \times C$, where $N = H \times W$. Briefly, given the input $X \in \mathbb{R}^{N \times C}$, the initial bases $\mu \in \mathbb{R}^{K \times C}$ and $Z \in \mathbb{R}^{N \times N}$ are the latent variables. The E-step is used to estimates the latent variables $Z$, and then used the M-step updated bases $\mu$. After $T$ times iteration, we reconstruct the $\hat{X}$ since $K << N$, $\hat{X}$ lies in a subspace of $X$. This method removes much unnecessary noise and makes the final classification of each pixel more segmentable. Moreover, this operation reduces the complexity from $O(N^2)$ to $O(NK)$ in the pixel segmentation process. 

E-step:
\begin{center}
	\begin{equation}
	z_{nk}=\frac{\kappa(x_n,\mu_k)}{\sum_{j=1}^{K}\kappa(x_n,\mu_j)}
	\end{equation}
\end{center}
where $\kappa $ represents the general kernel function, we simply take the exponential inner dot $exp(a^T,b)$ in our implementation.

M-step:
\begin{center}
	\begin{equation}
	\mu_k^t=\frac{z_{nk}^tX_n}{\sum_{m=1}^{N}z_{mk}^t}
	\end{equation}
\end{center}

\begin{figure*}[htb]
	\centering{\includegraphics[scale=0.25]{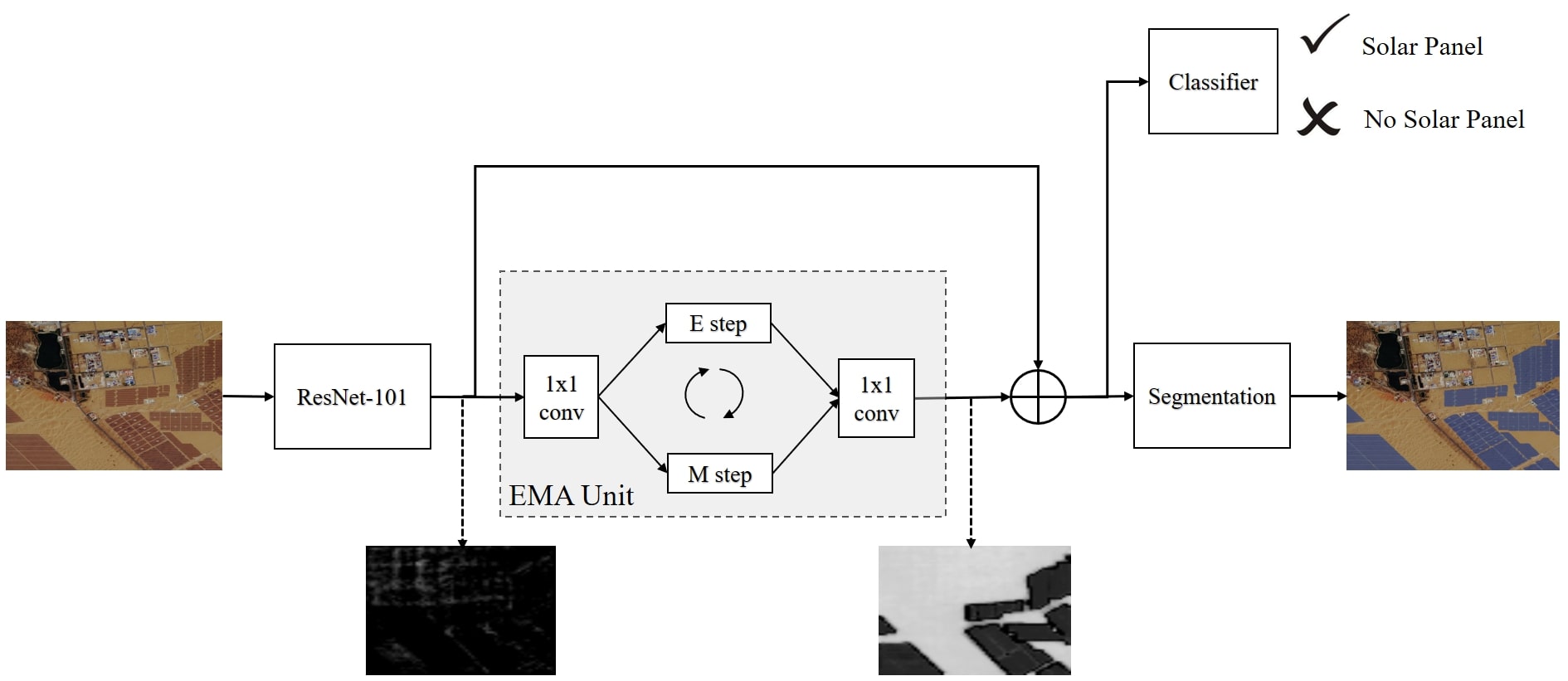}}
	\caption{SolarNet Architecture: in addition to the EMA operator, two $1\times 1$ convolutions at the beginning and the end of EMA, sum the output with original input, to form a residual-like block.}
	\label{fig:solarnet_art}
\end{figure*}

One shortcoming of  FCN segmentation structure is that its multiple local convolution operations is not  able to capture sufficient global information,and thus harms the performance in discontinuous object segmentation.  The structure of EMAU  based on EM algorithm is an unsupervised clustering algorithm without convolution operation and thus could effectively captures the global information. In our case, the solar power plants usually scatter in various discontinuous areas as shown in Figure \ref{fig:solarnet_art}, and EMANet is able to deal with such case as shown in the result section.

\begin{figure*}[htb]
	\centering{\includegraphics[scale=0.25]{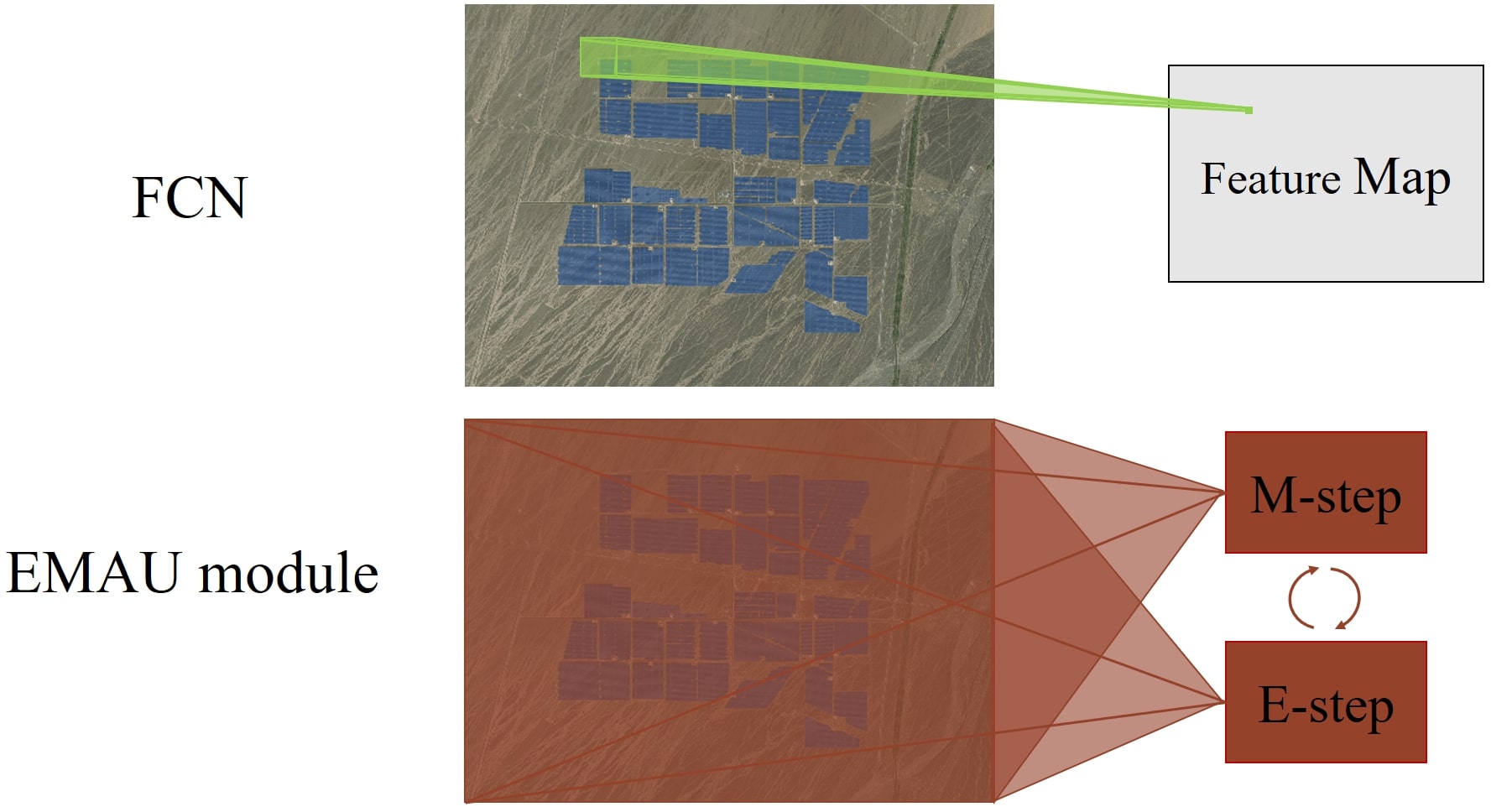}}
	\caption{When performing convolution operation, each convolution operator only extracts the local spatial features. By contrast, after  multi-level convolution operation, the continuous spatial information of the feature map is split by each convolution operator. The EMAU module performs clustering operation of element wise, and could capture more the global information in space.}
	\label{fig:emau}
\end{figure*}

Inspired by Zhou and Le's work\cite{zhou2016learning,le2019multitask}, we proposed an optimized multitask-EMANet, which combines local pixel-level segmentation and global image-level classification. Many existing studies show that the feature map of classification network usually corresponds the area of the object to be segmented, which could improve the segmentation performance. 

\begin{center}
	\begin{equation}
	{Loss}_{total} = \lambda\cdot{Loss}_{cls} + (1-\lambda)\cdot{Loss}_{seg}
	\end{equation}
\end{center}

Moreover, the work of DeepSolar\cite{yu2018deepsolar} did not  use the segmentation network but leveraged  the intermediate results from the classification branch and generated the Class Activation Maps (CAMs) by aggregating feature maps learned through the convolutional layers. This method did not require segmentation ground truth to train the model, but required the ground truth of class label to minimize the classification error.  

The proposed SolarNet architecture used pretrained ResNet-101 as backbone\cite{he2016identity} and the EMAU module to extract features. After re-configuring the features of EMAU module, the feature of ResNet-101 were then summed together and the last summed one was used to the last segmentation task.  SolarNet adopted the classification network to further enhance the segmentation results. Meanwhile, the classification network shares the same weight with segmentation network, and the final layer is a fully connected layer which is used to classify whether contains the solar planes or not. With single forward pass we then computed the segmentation loss and classification loss simultaneously. The network architecture is shown in Figure \ref{fig:solarnet_art}.

When training the model, we also adopted adam gradient descent method\cite{burges2005learning,dozat2016incorporating}. In order to fully incorporate the EMAU's into deep neural networks, we here describe how to train EMAU in each iteration. As each image $X$ has different pixel feature distributions compared to others, using the $\mu$ to reconstruct feature maps of a new image is not suitable. So we need to run EMAU moudle on each image simultaneously. For the first mini-batch, the Kaiming’s initialization\cite{he2015delving} has been used to initialize $\mu_0$ , where the matrix multiplication can be treadted as a $1\times 1$ convolution. For the following batches, we can simple used back propagation to update $\mu_0$ by standard. However, since iterations of E-step and M-step can be expanded as a recurrent neural network (RNN) \cite{mikolov2010recurrent}, the gradients propagating though them will generate the vanishing or explosion problem. Therefore, the updating of $\mu_0$ is unstable, moving averaging\cite{dandawate1995asymptotic} has been used to update $\mu_0$ in the training process. After several iterations over an image, the generated $\mu_T$ can be considered as a biased update of $\mu_0$, where the bias comes from the image sampling process. 
\begin{center}
	\begin{equation}
	\mu_0 = \alpha\mu_0 + (1-\alpha)\mu_T
	\end{equation}
\end{center}

The pseudo code of the training process of SolarNet is shown in Algorithm \ref{alg_train}. It is important to note that in each iteration  a semi-supervised clustering process of T-round EMAU module is required. And in the test process, each image was performed a clustering process with T-round iteration.

\renewcommand{\algorithmicrequire}{\textbf{Initial:}}  % Use Input in the format of Algorithm
\renewcommand{\algorithmicensure}{\textbf{Procedure:}} % Use Output in the format of Algorithm
\algnewcommand\algorithmicinput{\textbf{Input:}}
\algnewcommand\Input{\item[\algorithmicinput]}
\begin{algorithm}[htb]
	\caption{SolarNet Training Procedure}
	\begin{algorithmic}[1]
		\Require
		
		Random Initial network's weights: $W_0$
		\Input
		
		Original Satellite Imagery: $X$
		
		Semantic Segmentation Imagery: $X_s$
		
		Whether it contains solar panels: $Y$
		\Ensure
		\Function {EStep}{$\mu,X$}
		\State \Return{$z_{nk}=\frac{\kappa(x_n,\mu_k)}{\sum_{j=1}^{K}\kappa(x_n,\mu_j)}$}
		\EndFunction
		\Function {MStep}{$Z,X$}
		\State \Return{$\mu_k^t=\frac{z_{nk}^tX_n}{\sum_{m=1}^{N}z_{mk}^t}$}
		\EndFunction
		\For{$i=0 \to Max Iter$}
		\State $X_{res} = ResNet(X)$
		\State $Logit = Cls(X_{res})$
		\State $L_{cls} = CrossEntropy(Logit,Y)$
		\State Random initial $\mu_0$
		\State $Z_0 = \Call{EStep}{\mu_0,X_{res}}$
		\For{$t=0 \to T$}
		\State $u_t = \Call{MStep}{Z_t,X_{res}}$
		\State $Z_{t+1} = \Call{EStep}{\mu_t,X_{res}}$
		\EndFor
		\State $\widetilde{X} = Z_t \cdot \mu_t$
		\State $L_{seg} = CrossEntropy(\widetilde{X},X_s)$
		\State ${L}_{total} = \lambda\cdot{L}_{cls} + (1-\lambda)\cdot{L}_{seg}$
		\State $W_{i+1} = W_i + \frac{\partial {L}_{total}}{\partial W_i}$
		\EndFor
	\end{algorithmic}
	\label{alg_train}
\end{algorithm}

\section{Results}
In this section, we elaborated the implementation details of SolarNet and demonstrated the results of all the solar farms in China that we have mapped. First we compared the performance of SolarNet and  two other baseline methods with regard to three kinds of datasets. Then we visualized the locations and  distributions of  all  solar power plants in China detected by SolarNet. Furthermore, we showed several bad cases and discussed how to future improve our algorithms in the future. 

819 images were used to train the mode  while  119 images were used to test the model.  The size of all the images  ranges from  $512\times 512$ to $10000\times 10000$. In order the create more dataste to train the model, we adopted the following data augmentation methods: 
\begin{itemize}
	\item [.] 
	\textbf{Crop}: Choosed a random ROI area from a original image: $X_{arg}=ROI(X)$.       
	\item [.]
	\textbf{Scale}: Choosed a random scale size $s \in (0.8,1.2)$, rescaled the original image: $X_{arg}=Rescale(X,s)$
	\item [.]
	\textbf{Rotation}: Choosed a random angle $\theta \in (-180,180)$, rotated the orignal image:$X_{arg}=Rotate(X,\theta)$
	\item [.]
	\textbf{Reflection}: Flipped the original image horizontally: $X_{arg}=FlipH(X)$, or flipped the original image vertically: $X_{arg}=FlipV(X)$
\end{itemize}

\begin{table}[H]
	\centering
	\begin{tabular}{|c|c|c|c|c|}
		\hline
		\textbf{Parameter} &  Learning Rate & Iteration & Training Set & Testing Set\\
		\hline
		\textbf{Value} & $1e^{-3}$ & $20000$ & $819$ & $119$ \\
		\hline
		\multicolumn{1}{|c|}{\textbf{Parameter}} & \multicolumn{2}{|c|}{EM Iteration} & \multicolumn{2}{|c|}{EM Latent Variables Size}\\
		\hline
		\multicolumn{1}{|c|}{\textbf{Value}} & \multicolumn{2}{|c|}{$10$} & \multicolumn{2}{|c|}{$1024$}\\
		\hline
	\end{tabular}
	\caption{ Parameters of SolarNet to train the model.}
	\label{fig:param}
\end{table}

\begin{figure}[H]
	\centering
	\subfigure[ORIGINAL IMAGERY]{
		\begin{minipage}[b]{0.2\linewidth}
			\includegraphics[width=1\linewidth]{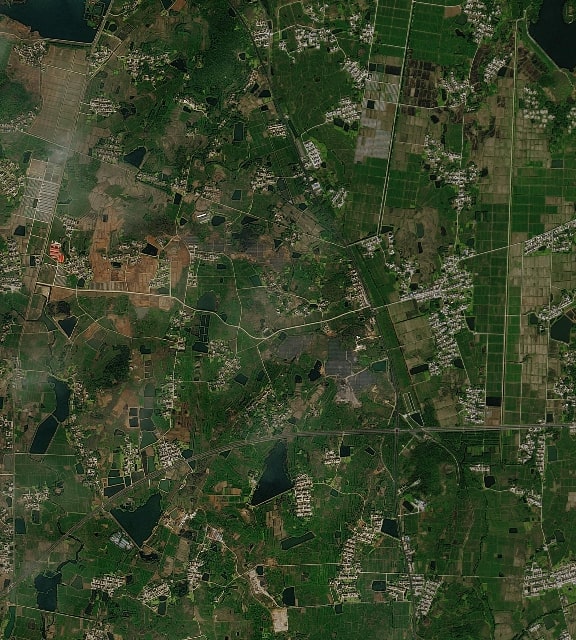}\vspace{4pt}
			\includegraphics[width=1\linewidth]{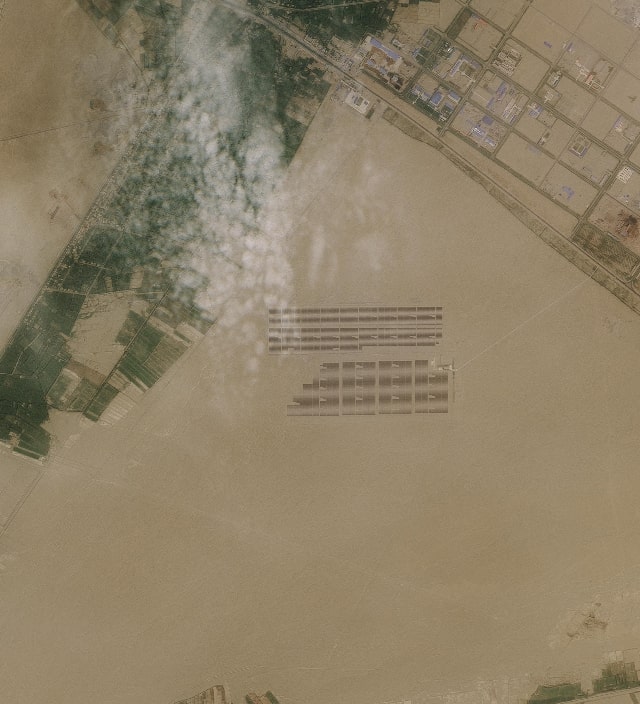}\vspace{4pt}
			\includegraphics[width=1\linewidth]{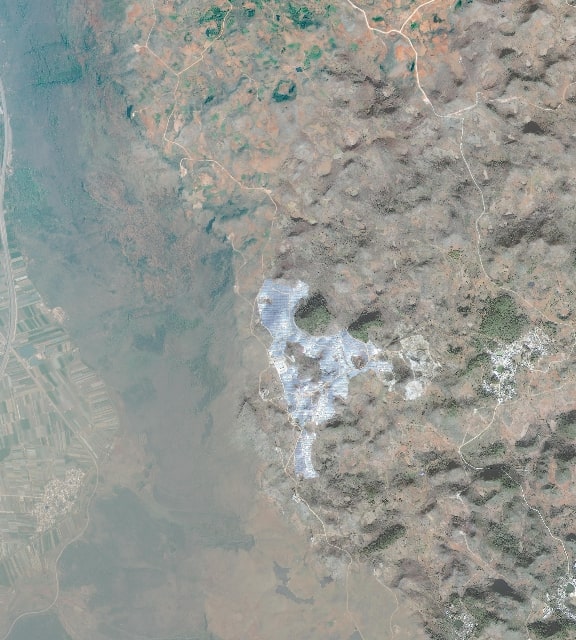}\vspace{4pt}
			\includegraphics[width=1\linewidth]{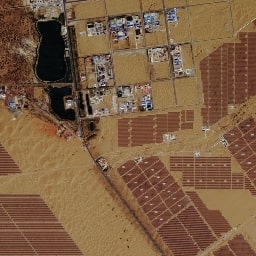}
	\end{minipage}}
	\subfigure[UNET SEGMENTATION]{
		\begin{minipage}[b]{0.2\linewidth}
			\includegraphics[width=1\linewidth]{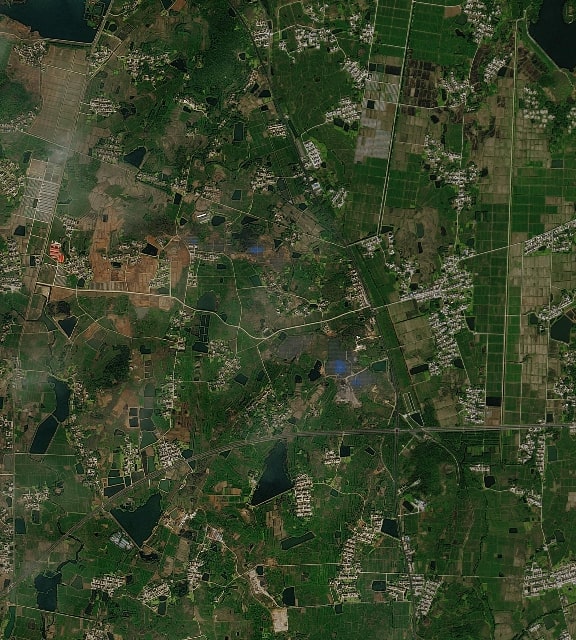}\vspace{4pt}
			\includegraphics[width=1\linewidth]{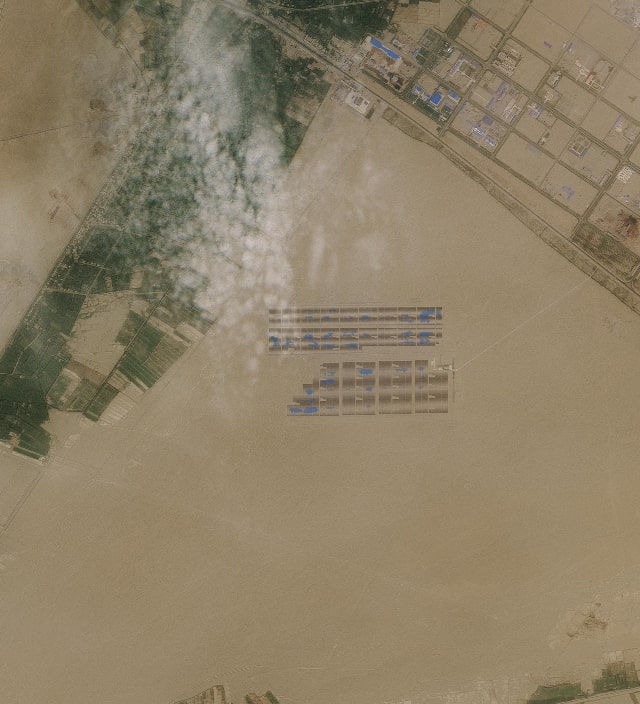}\vspace{4pt}
			\includegraphics[width=1\linewidth]{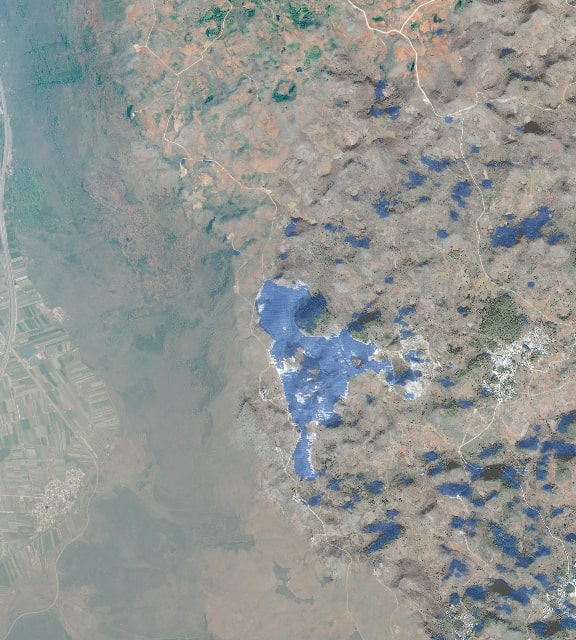}\vspace{4pt}
			\includegraphics[width=1\linewidth]{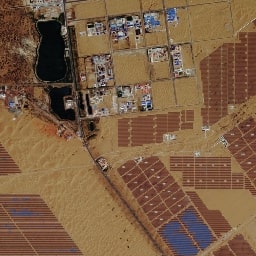}
	\end{minipage}}
	\subfigure[SOLARNET SEGMENTATION]{
		\begin{minipage}[b]{0.2\linewidth}
			\includegraphics[width=1\linewidth]{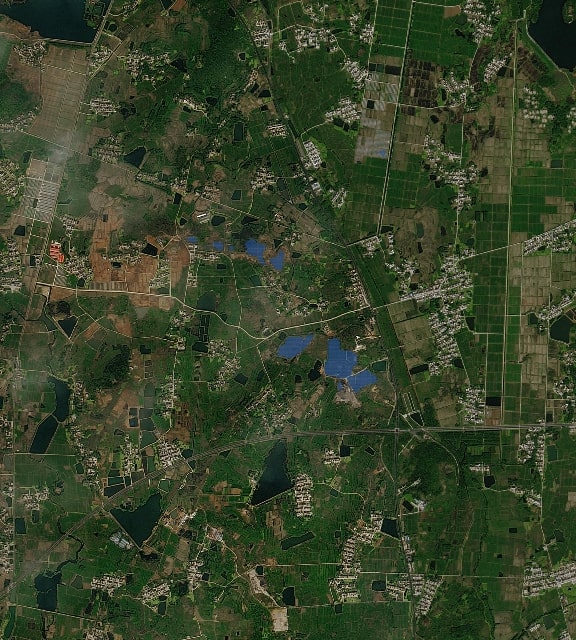}\vspace{4pt}
			\includegraphics[width=1\linewidth]{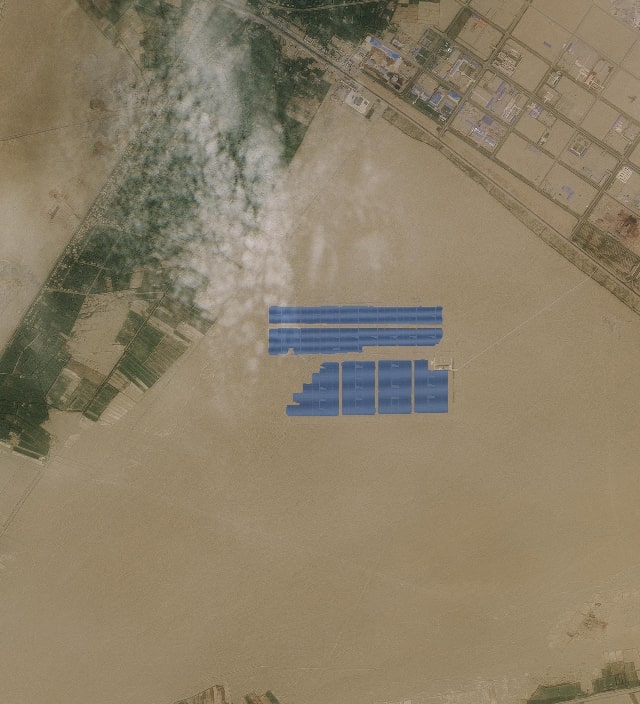}\vspace{4pt}
			\includegraphics[width=1\linewidth]{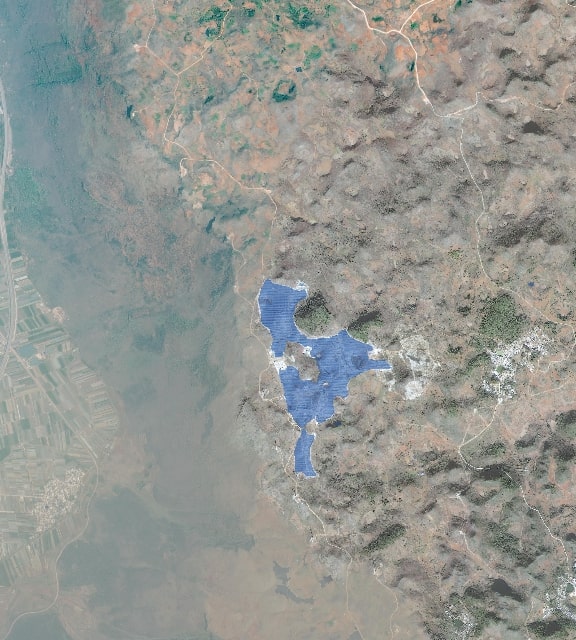}\vspace{4pt}
			\includegraphics[width=1\linewidth]{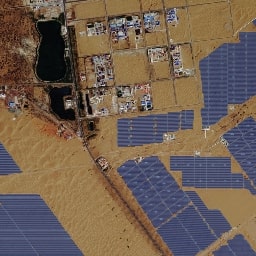}
	\end{minipage}}
	\subfigure[GROUND TRUTH]{
		\begin{minipage}[b]{0.2\linewidth}
			\includegraphics[width=1\linewidth]{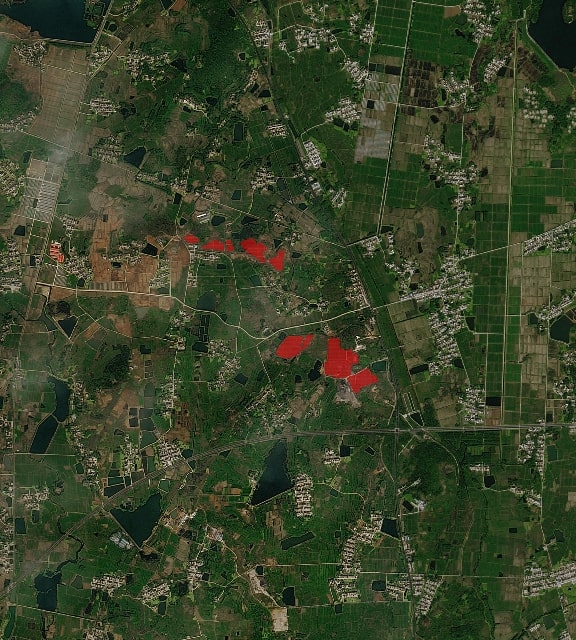}\vspace{4pt}
			\includegraphics[width=1\linewidth]{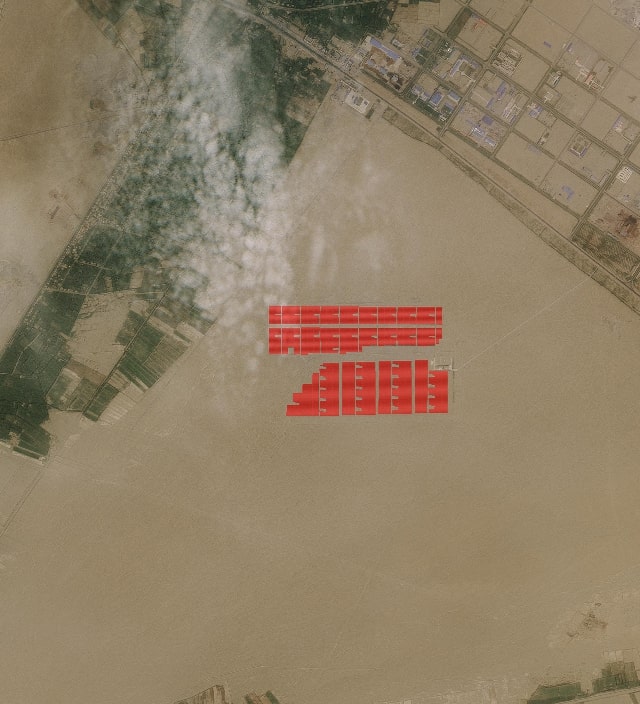}\vspace{4pt}
			\includegraphics[width=1\linewidth]{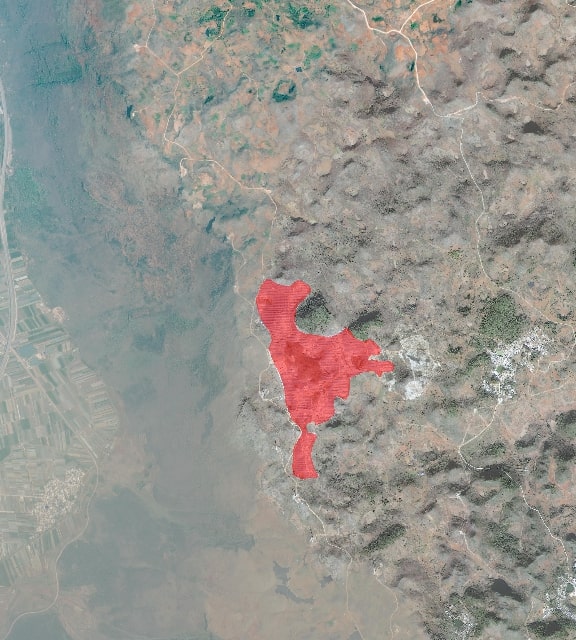}\vspace{4pt}
			\includegraphics[width=1\linewidth]{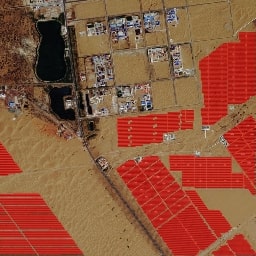}
	\end{minipage}}
	\caption{Solar farms located by SolarNet. The first column is the orignal satellite imagery data. The blue area indicates the detected solar farms by UNet (second column) and EMANet (third column) and red area in the fourth column indicates the ground-truth labeled manually. One can see how SolarNet was able to accurately detect solar farms under very complicated backgrounds.}.
	\label{tab:visual_case}
\end{figure}

We used mean Intersection over Union (mIoU)  as the criteria to evaluate segmentation performance and compared the SolarNet with two other methods. The results in Table \ref{tab:eval_ret} shows that the SolarNet outperformed two others.  Figure \ref{tab:visual_case} demonstrated several solar farms detected by all three methods and one can see that SolarNet is able to accurate detect the solar farms under very complex backgrounds.  Figure \label{fig:pand_horse}  showed two sizeable solar farms we detected which shaped like a horse and panda, respectively. 

\begin{table}[H]
	\centering
	\begin{tabular}{|c|c|c|c|}
		\hline
		\multicolumn{1}{|c|}{\textbf{Model}} & \multicolumn{3}{|c|}{\textbf{mIOU}} \\
		\hline
		~ & our dataset & deepsolar dataset & our+deepsolar dataset\\
		\cline{2-4}
		Resnet101-Unet & 84.65\% & 84.22\% & 86.54\% \\
		\hline
		Resnet101-EMANet-single & 94.00\% & \textbf{90.98}\% & 93.79\% \\
		%\hline
		%SolarNet-Multitask-0.999 & 93.87\% & 90.52\% & 93.64\% \\
		\hline
		SolarNet-Multitask-1.0 & \textbf{94.21}\% & 90.39\% & \textbf{93.94}\% \\
		\hline
	\end{tabular}
	\caption{ With the multi-task embedding, SolarNet could beat the orignal EMANet and UNET on our dataset evaluation.}
	\label{tab:eval_ret}
\end{table}

%\begin{comment}
% draw example pictures
%\end{comment}

\begin{figure}[H]
	\centering
	\subfigure[ORIGINAL IMAGERY]{
		\begin{minipage}[b]{0.45\linewidth}
			\includegraphics[width=1\linewidth]{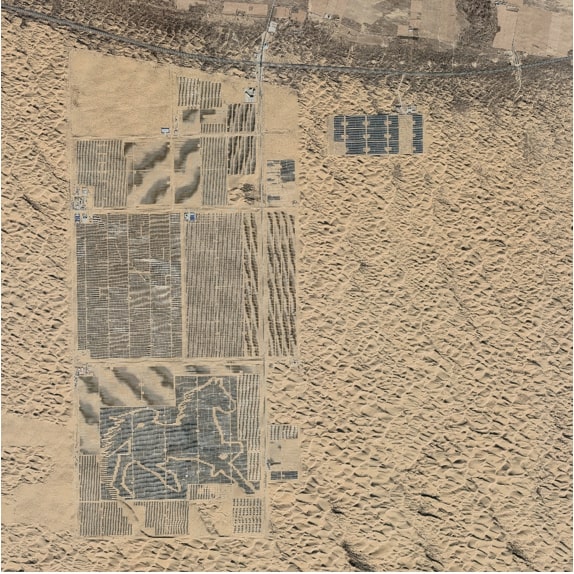}\vspace{4pt}
			\includegraphics[width=1\linewidth]{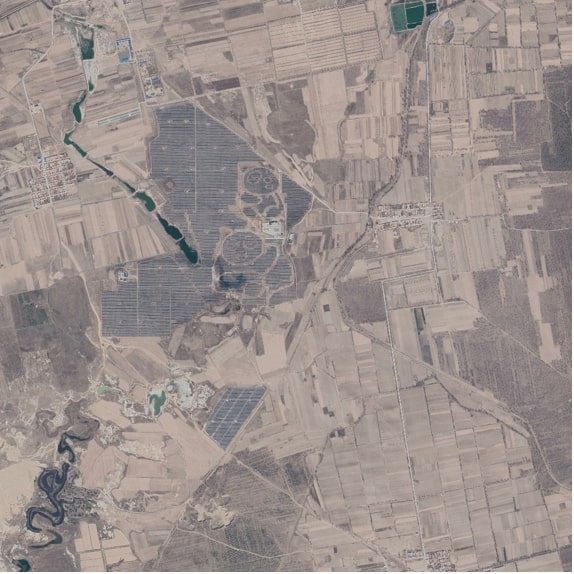}
	\end{minipage}}
	\subfigure[SOLARNET SEGMENTATION]{
		\begin{minipage}[b]{0.45\linewidth}
			\includegraphics[width=1\linewidth]{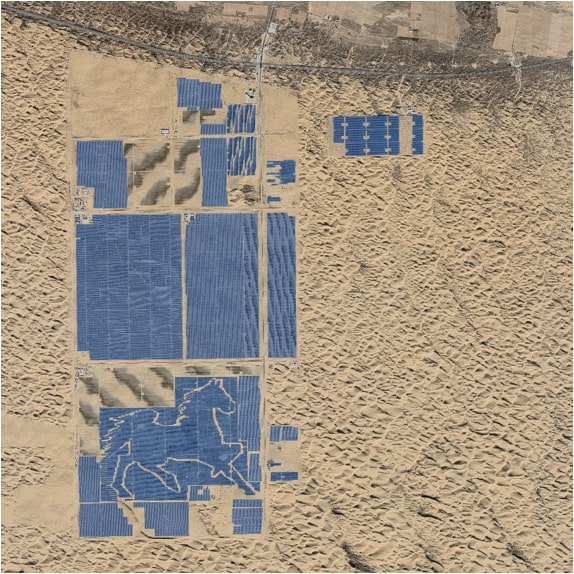}\vspace{4pt}
			\includegraphics[width=1\linewidth]{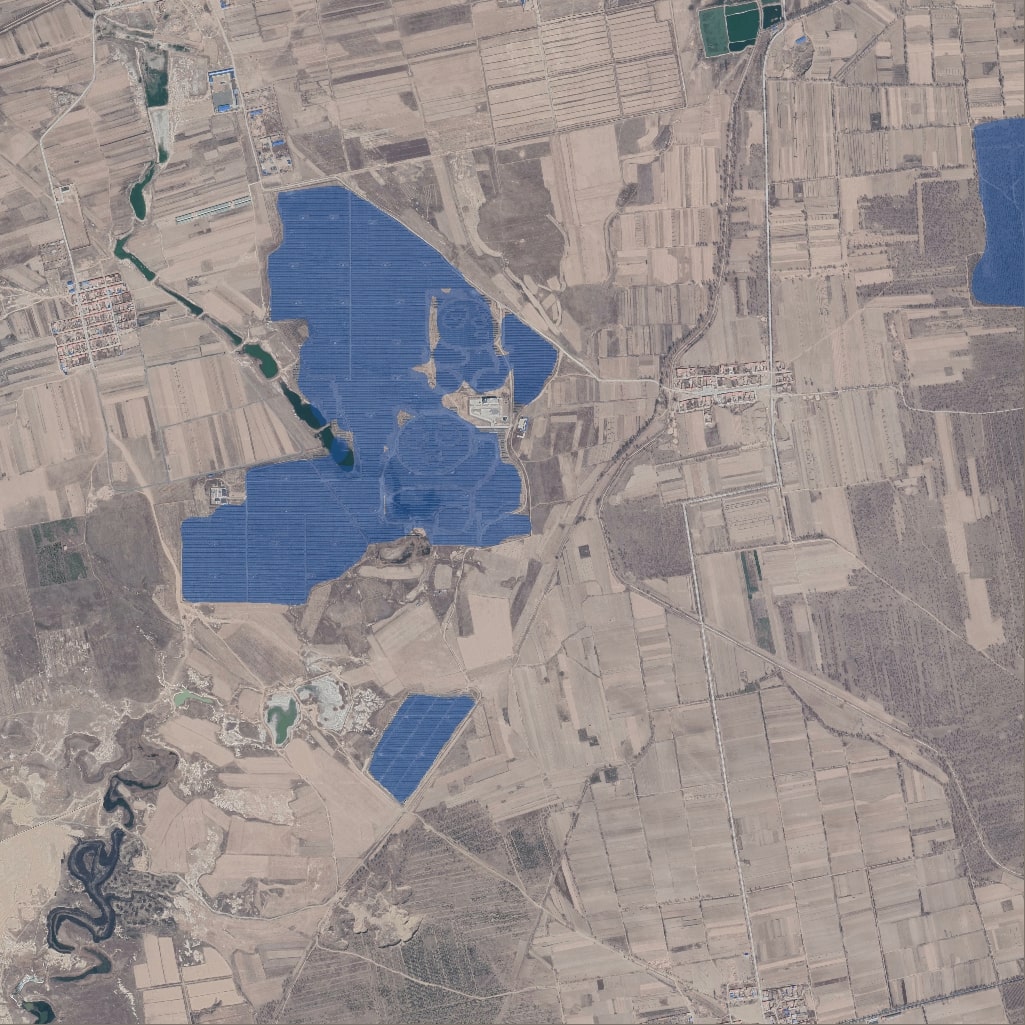}
	\end{minipage}}
	\caption{Two massive animal-shaped (horse and panda) solar farms detected by SolarNet.}
	\label{fig:pand_horse}
\end{figure}

\begin{figure}[H]
	\begin{center}
		\includegraphics[scale=0.3]{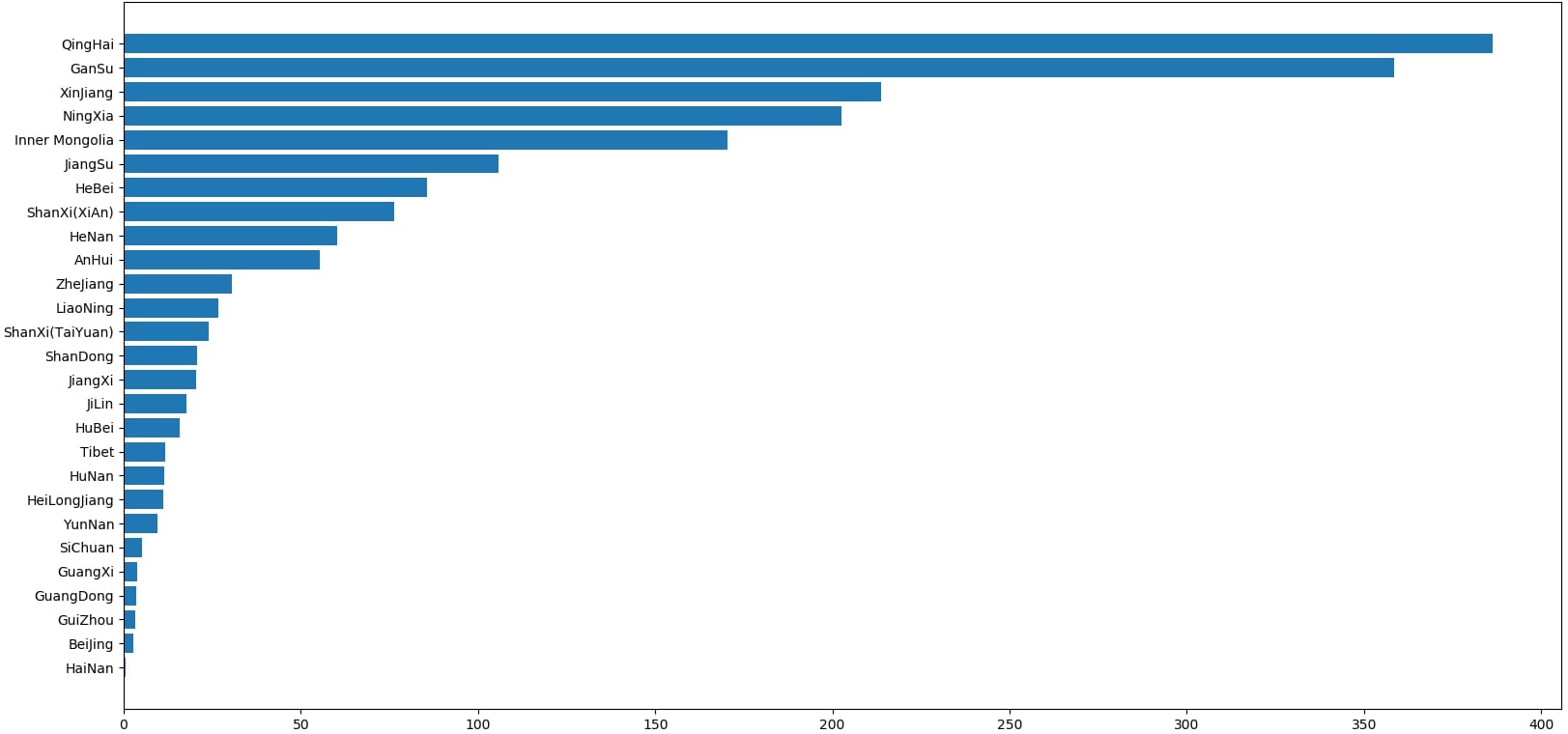}	
	\end{center}
	\caption{The area of detected solar farms in various provinces in China (unit: $km^2$)}
	\label{fig:province1}
\end{figure}

\begin{figure}[H]
	\begin{center}
		\includegraphics[scale=0.4]{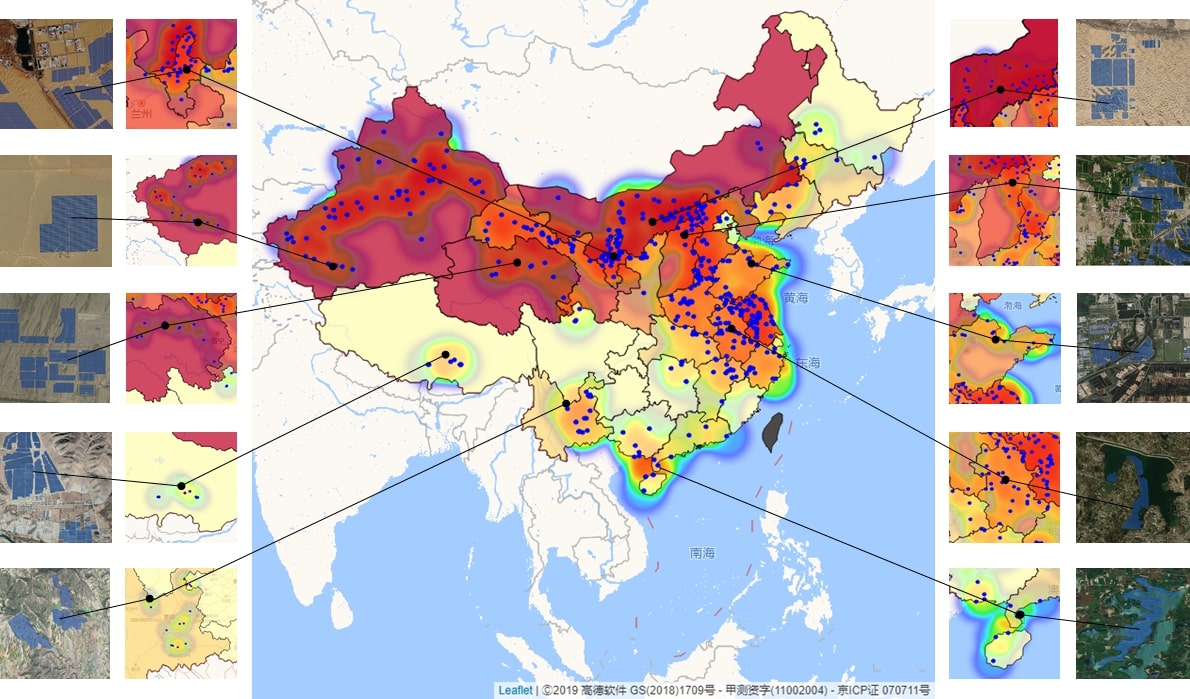}
	\end{center}
	\caption{ Solar farm map in China. Each blue dot indicates a detected solar farm from satellite imagery. We colored each province according to the area of solar farms (darker color indicates larger areas). A heat map of solar farm density was also overlaid. 
	Ten representative solar farms built on deserts, mountains, lakes or the fields were also displayed.}
	\label{fig:province2}
\end{figure}

We then used the trained SolarNet framework to map all the solar farms in China by mining large scale satellite imagery data that covered the whole China. We successfully  detected about 500 solar farms covering the area of 2000 square kilometers or 770 square miles in total, equivalent to the size of whole Shenzhen city or two and a half or New York city. Figure \ref{fig:province2} visualized the locations of all detected solar farms in China marked by blue dots. One can see that most of the solar farms were built in the northwestern part of China where the sunlight is abundant and thus is ideal for solar power. Among all the provinces in China, Qinghai has installed the most solar farms with the area of near 400 square kilometers in total as shown in Figure \ref{fig:province1}.

\section{Discussion and future work}
In this paper, we proposed a deep learning framework named SolarNet to map the solar farms from massive satellite imagery data. The method was also evaluated by comparing with two other image segmentation algorithms and the results showed the accuracy of SolarNet. We then used SolarNet to successfully detect near 500 large solar farms in China, covering near 2000 square kilometers equivalent to the whole size of Shenzhen city. To the best of our knowledge, it is the first time that we identified the locations and sizes of solar farms on satellite imagery through deep learning in China, the largest producer of solar power in the world.

\begin{figure}[H]
	\centering
	\subfigure[ORIGINAL IMAGERY]{
		\begin{minipage}[b]{0.3\linewidth}
			\includegraphics[width=1\linewidth]{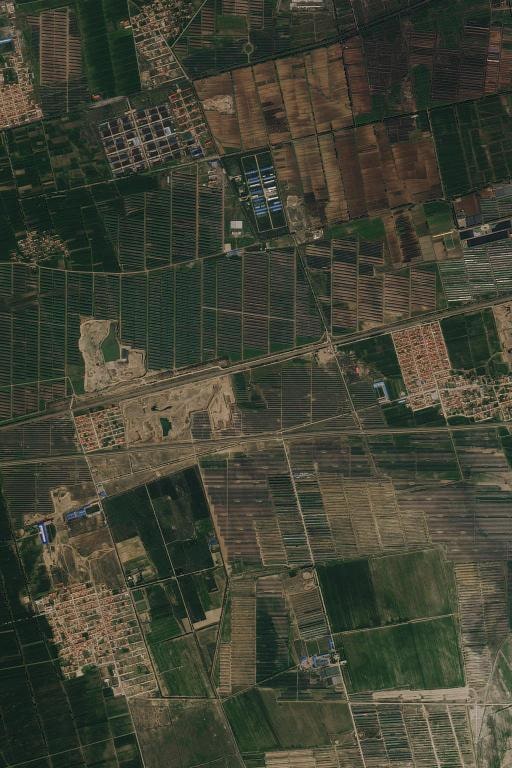}\vspace{4pt}
			\includegraphics[width=1\linewidth]{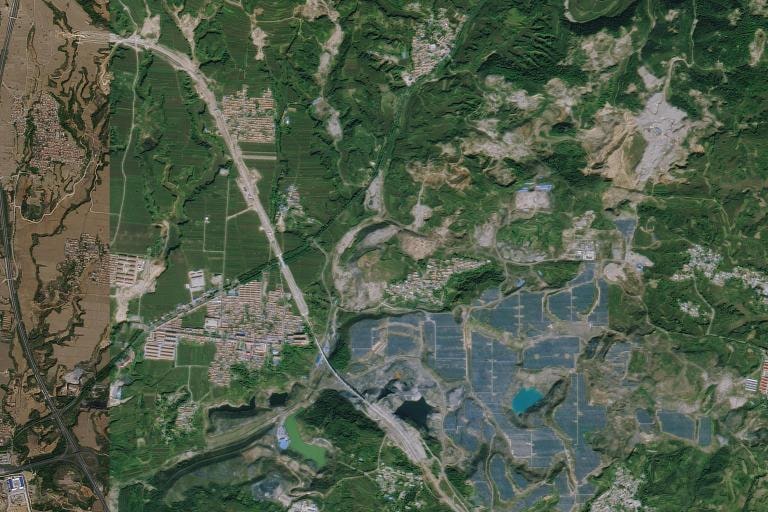}
	\end{minipage}}
	\subfigure[SOLARNET SEGMENTATION]{
		\begin{minipage}[b]{0.3\linewidth}
			\includegraphics[width=1\linewidth]{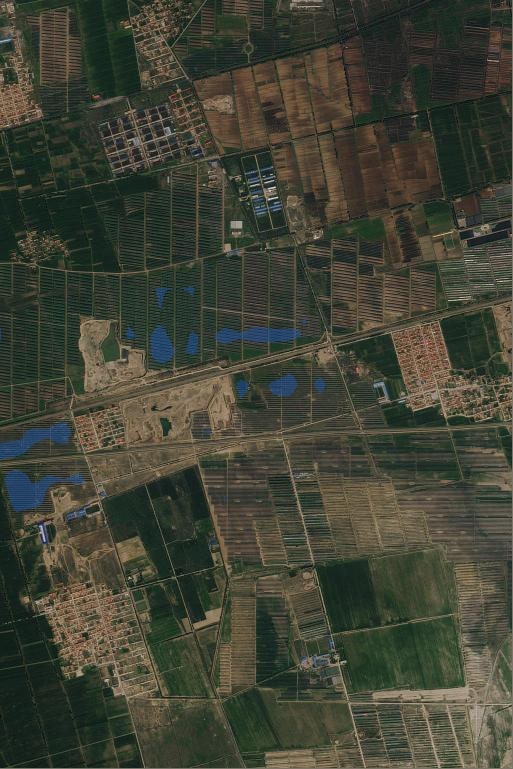}\vspace{4pt}
			\includegraphics[width=1\linewidth]{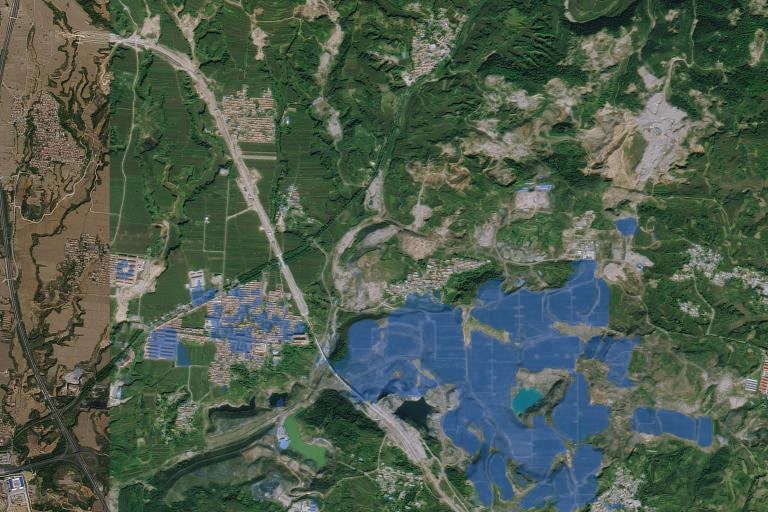}
	\end{minipage}}
	\subfigure[GROUND TRUTH]{
		\begin{minipage}[b]{0.3\linewidth}
			\includegraphics[width=1\linewidth]{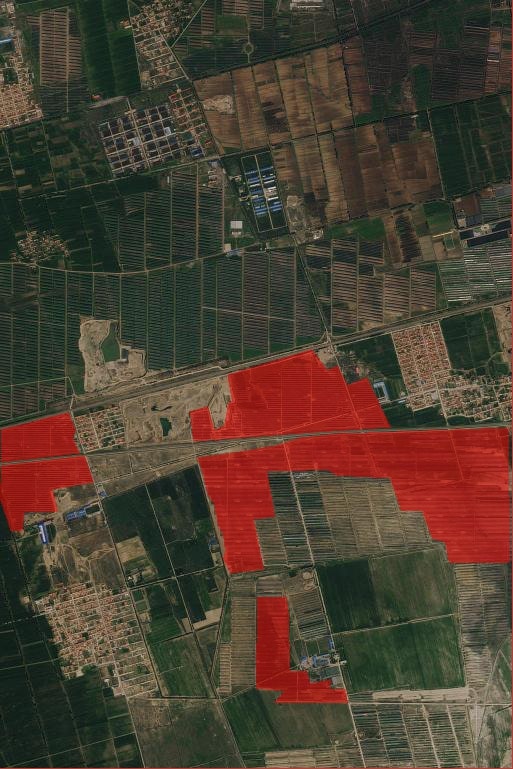}\vspace{4pt}
			\includegraphics[width=1\linewidth]{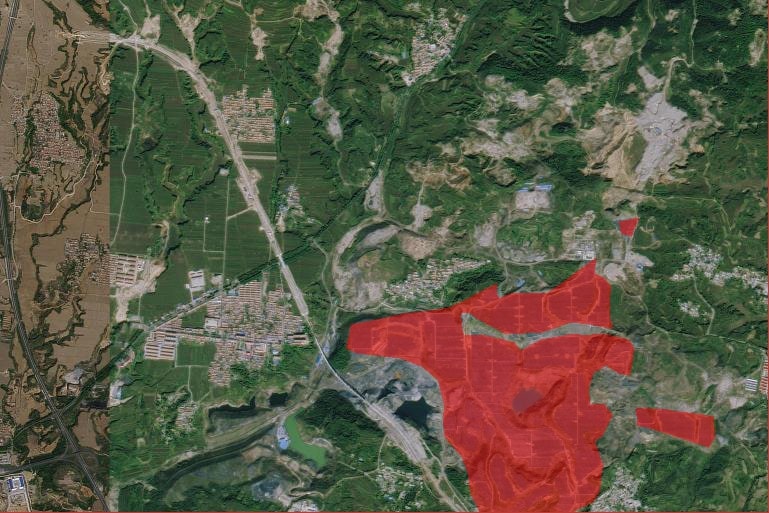}
	\end{minipage}}
	\caption{SolarNet may fail to detect the solar farm when it resembles its surrounding environment.}
	\label{fig:badcase}
\end{figure}

SolarNet may fail to detect the solar farms when the it resembles its surrounding background as shown in Figure \ref{fig:badcase}. In the future, we plan to improve our methods in the following way:
\begin{itemize}
	\item [1)] 
Labeling more solar panels from the satellite imagery data in various circumstances, such as the solar panels on the roof in residential areas. 
	\item [2)]
Adapting SolarNet to handle the satellite imagery data with various resolutions \cite{benz2004multi}. For example, HRNet proposed by \cite{sun2019deep}  is an effective super-resolution method to deal with various resolution images.
	\item [3)]
Using hyperspectral imagery data to enhance the segmentation performance.  As showed in \cite{alexakis2009detection,arellano2015detecting} could provide more information when detecting objects from satellite.
\end{itemize}

Mapping and tracking the installment of solar panel from satellite imagery data is very helpful for the following fields: 1) it could help the solar PV  power companies to optimize the location and direction of solar panels so that they can maximize their renewable energy production; 2) it could help the investors and market researchers to track the latest trends of solar power industry; 3) the government could evaluate their policy efficiency based on our results, for example, how the  subsidiary policy is impacting the development of solar power industry.  Therefore, we plan to build a Solar Power Index in China by analyzing longer historical satellite imagery data with SolarNet so that we could track long term trends. And we also plan to apply the proposed framework to map the locations and develop the index of other type of renewable energy such as wind turbine.

\bibliographystyle{IEEEtran}
% argument is your BibTeX string definitions and bibliography database(s)
\bibliography{IEEEabrv,bibfile}

\end{document}